**Motility at the origin of life: Its characterization and a model**[1]

Tom Froese [a, b, c *], Nathaniel Virgo [d, c], and Takashi Ikegami [c]

[a] Departamento de Ciencias de la Computación
Instituto de Investigaciones en Matemáticas Aplicadas y en Sistemas
Universidad Nacional Autónoma de México, DF, Mexico

[b] Centro de Ciencias de la Complejidad
Universidad Nacional Autónoma de México, DF, Mexico

[c] Ikegami Laboratory
Department of General Systems Studies
University of Tokyo, Tokyo, Japan

[d] Max Planck Research Group on Biospheric Theory and Modelling
Max Planck Institute for Biogeochemistry, Jena, Germany

* Corresponding author: t.froese@gmail.com

---

**Abstract**

Due to recent advances in synthetic biology and artificial life, the origin of life is currently a hot topic of research. We review the literature and argue that the two traditionally competing 'replicator-first' and 'metabolism-first' approaches are merging into one integrated theory of individuation and evolution. We contribute to the maturation of this more inclusive approach by highlighting some problematic assumptions that still lead to an impoverished conception of the phenomenon of life. In particular, we argue that the new consensus has so far failed to consider the relevance of intermediate timescales. We propose that an adequate theory of life must account for the fact that all living beings are situated in at least four distinct timescales, which are typically associated with metabolism, motility, development, and evolution. On this view, self-movement, adaptive behavior and morphological changes could have already been present at the origin of life. In order to illustrate this possibility we analyze a minimal model of life-like phenomena, namely of precarious, individuated, dissipative structures that can be found in simple reaction-diffusion systems. Based on our analysis we suggest that processes in intermediate timescales could have already been operative in prebiotic systems. They may have facilitated and constrained changes occurring in the faster- and slower-paced timescales of chemical self-individuation and evolution by natural selection, respectively.

# 1. Introduction

In the field of artificial life and synthetic biology there is a widespread optimism that the creation of a living cell from scratch is imminent [14, 35, 98, 112]. Accordingly, it is hoped that these bio-engineering approaches will help to finally resolve one of the most important outstanding mysteries of science, namely the problem of the origin of life on earth. In this paper we highlight and criticize some unquestioned assumptions underlying these approaches, and offer some ideas on how to improve them.

In the science of the origin of life there are two competing traditions, which are called 'replicator-first' (a.k.a. 'information-first) and 'metabolism-first', with the 'replicator-first' scenario generally receiving most of the support [1, 74]. We briefly review the main tenets of these two traditions, and argue that they are beginning to merge into one mainstream consensus, which we call the 'information-compartment-metabolism-first' approach to the origin of life. In essence, the consensus is that the crucial event in the first transition from non-living matter to a living being is the appearance of a bounded self-maintaining entity subject to evolution by natural selection. Researchers' opinions mainly differ in terms of the nature of the heritable information of the first replicators, i.e. whether replication involved symbolic representations and/or metabolic structures. However, this is a difference in emphasis rather than an absolute distinction. No matter whether the first forms of life are hypothesized to have been based on strings of RNA code or self-maintaining vesicles, the underlying shared assumption can be summed up by the slogan that 'life began when evolution began' [101]. Accordingly, most research on the origin of life is focused on how best to simulate or chemically engineer the emergence of self-replicating structures that are capable of open-ended evolution [26, 77, 78, 94].

The information-compartment-metabolism-first consensus is an integrated attempt to understand the requirements for evolvability, individuation, and self-organization, all of which are characteristic features of life [84, 85]. In this respect the consensus is to be welcomed over the more limited traditional perspectives of the replicator-first and metabolism-first approaches. Nevertheless, we argue that the current focus on spatial encapsulation in the service of both self-maintenance and natural selection remains an impoverished view of the phenomenon of life. Because the consensus only takes the two disparate timescales of chemical reactions and evolutionary history into account, it has failed to consider the relevance of the intermediate timescales of behavior and development. In other words, this approach completely neglects what is perhaps most characteristic of living beings in contrast to non-living things, namely self-generated translational and transformative movements in response to internal and environmental events. All living beings engage in interactive exchanges with their environment, including with other living beings, and they also modify their morphological structure over time. We propose that such activities in these intermediate timescales are not limited to more recently evolved forms of life; they may have already been operative at the very origin of life.

In order to illustrate this idea we analyze a minimal simulation model of a simple class of dissipative structures, namely a reaction-diffusion system. We follow Virgo [107] in arguing that dissipative structures whose self-production is spatiotemporally localized, but not membrane-bound, have many important commonalities with living beings. They are therefore worthy of study in the context of the origin of life. The model demonstrates that even very simple examples of such self-producing structures are capable of motility, adaptive behavior, structural modification, and epigenetic evolution. On the basis of these modeling results we conclude by speculating about the feasibility of a 'movement-first' approach to the origin of life.

## 2. Current theories about the origin of life

2.1 The replicator-first approach

The replicator-first approach to the origin of life assumes that there was already a genetic code right at the start of life itself. An extreme version of this view is known as the 'RNA world', which holds that "the first stage of evolution proceeds […] by RNA molecules performing the catalytic activities necessary to assemble themselves from a nucleotide soup" [36, p. 618]. However, it is now being considered that this traditional RNA-only view is incomplete. Maynard-Smith was the first to argue that the evolution of a system of autocatalysts would be vulnerable to the emergence of parasitic side-reactions and would require an unlikely kind of group selection of all the individual chemical species [59]. He also suggested that both issues could be addressed by assuming that the chemical system was compartmentalized inside some kind of membrane. On this view, compartments turn what would otherwise be a cluttered 'ecosystem' of competing genes into a population of distinct genotypes, thereby satisfying a crucial tenet of the neo-Darwinian theory of evolution [45].

The replicator-first approach has therefore turned toward the task of incorporating suitable information-encoding molecules into the right kind of vesicle in a way that ensures the accurate replication of both [35, 38, 88]. Of particular importance is to find a way of encapsulating the genetic code such that it satisfies another requirement of Darwinian evolution, namely competition and differential success [11, 12]. Although the growing emphasis on the need for an enclosing membrane at the origin of life brings this view closer to the metabolism-first approach, as we argue below, some differences remain. According to this replicator-first view, the role of metabolism for the origin of the first living cell is still mostly a secondary aspect, and it is occasionally absent from explicit consideration. On this view, the essence of life only consists of two complementary components: "fundamentally, a cell consists of a genome, which carries information, and a membrane, which separates the genome from the external environment" [11, p. 1558]. Thus, replicator-first theorists have come to generally accept the need for some kind of specialized self-individuation, even if it is often conceived like a passive container instead of an active membrane interface.

Further revisions to the replicator-first approach are underway. There is a growing recognition among proponents of the RNA world hypothesis that metabolic cycles played an essential role at the origin of life [22]. Even replicators would need some kind of metabolic capability "that would enable thermodynamic impediments on replicative capability to be largely circumvented" [75, p. 728]. Furthermore, the same metabolic mechanisms by which thermodynamic equilibrium is avoided can serve as a source of functional variation on which natural selection can potentially operate so as to increase diversity and complexity [111], at least as long as the variations are heritable. For instance, the famous evolutionary experiments by Spiegelman and Orgel demonstrated that RNA replication just for replication's sake, i.e. without any other fitness criterion than replication, does not lead to an increase in complexity [68]. On the contrary, less complex (shorter) RNA chains replicate faster, so in the absence of any other useful functionality, the RNA was selected to become shorter and shorter. What these experiments demonstrate is that the origins of life and the start of diversification via evolution require more than only the traditionally assumed triad of multiplication, variation, and heredity. Individuals need to be metabolically organized in a way that allows for the expressions of a variety of functions at the level of the living system as a whole for natural selection to properly get off the ground [64, 86].

Given these changes in emphasis it seems that the original replicator-first approach is developing into a more encompassing information-compartment-metabolism-first approach, where all the three molecular mechanisms are linked in an interdependent and heritable manner. An early example of this position is Ganti's [34] 'chemoton' theory, but there are also a growing number of modern theories that specifically highlight such a threefold design [79, 94].

2.2 The metabolism-first approach

Traditionally, the metabolism-first view has focused on more abstract issues regarding biological autonomy, self-organization, and self-production, such as Maturana and Varela's [58] autopoietic theory, Kauffman's [46] autocatalytic theory, and Rosen's [82] metabolic-repair system theory. It was widely assumed that living is essentially a process of self-production, which constitutes a spatially localized individual. Since this view of life does not necessitate the existence of information represented in a genetic code, it seems that life must have initially arisen because of constraints that were imposed by self-organization rather than natural selection. Nevertheless, more recent versions of metabolism-first approaches are no longer that different from the information-compartment-metabolism-first approach described above.

First, there is the idea of individuation, which is commonly inspired by the semi-permeable membrane of a living cell and is therefore often identified with a distinct physical boundary. For instance, according to Luisi and Varela's [53] interpretation of autopoietic theory, a self-producing network of autocatalytic chemical processes is not a sufficient condition for life, unless it is encapsulated within a physical boundary that enables the same processes which also produce the boundary itself. Similarly, modeling studies of autopoiesis tend to assume that a distinct physical boundary is necessary to prevent the network of processes from passively diffusing into the environment [10, 61, 105] and to regulate its molecular exchanges [8, 83]. On this view, the first living being was a bounded self-producing chemical system.

Second, there are efforts to integrate the emphasis on biological autonomy with the historical-collective requirements of evolution by natural selection [23, 84]. Research in prebiotic chemistry has demonstrated that it is possible to engineer the emergence of membrane-bounded micelles that are self-maintaining and self-replicating [2, 70, 109]. Radical proposals hold that natural selection is already operative at the chemical level, and that it can therefore guide the emergence of suitable autocatalytic systems [25, 46, 62]. On this view, heredity can be based on the chemical composition of the system itself, which serves as a 'compositional genome' [90]. Metabolism may also play a role in making novel functional variations available for selection [76]. Differential replicative success of the offspring is also ensured, because "Darwinian competitive exclusion is rooted in the chemical competitive exclusion of metabolism" [65, p. 58]. Indeed, simulation models have demonstrated that under some conditions the growth and division of membrane-bounded autocatalytic systems is sufficient for differential replicative success [69, 71]. This form of epigenetic evolution does not fulfill all of the requirements of the replicator-first approach, specifically 'unlimited heredity' based on modular templates [97], but it does involve natural selection in the classic Darwinian sense of multiplication, variation, and heredity.

As the evolutionary bottlenecks faced by protocells with composition-based genomes are clarified [106], there is a growing need for metabolism-first approaches to address the decoupling between phenotype and genotype that is realized in all extant living systems by means of a genetic code [86]. However, since genetic decoupling is not an all or nothing affair, with some molecules serving both functions, the metabolism-first view is on a continuum with the replicator-first approach. Both approaches can be integrated into one information-metabolism-compartment-first approach.

2.3  Toward an information-compartment-metabolism-first approach

The replicator-first and metabolism-first approaches to the origin of life have differed significantly in the past. However, due to recent advances they are no longer mutually exclusive. Both approaches generally agree that biological individuation requires a distinct physical compartment, even though they have different primary reasons for this claim (i.e. 'unit of natural selection' versus 'unit of self-production'). Both also agree that survival is about maintaining an adequate metabolism, although again their emphasis is different (i.e. 'self-replication' versus 'self-production'). In addition, they both acknowledge that the beginning of life was already shaped by natural selection, even though they disagree about the nature of the inherited information (i.e. 'modular template' or 'compositional genome'). In other words, while there are still differences between the replicator-first and metabolism-first approaches, they are merging into an integrated information-compartment-metabolism-first approach. This consensus about the origin of life is centered on the integration of phenomena that are taking place at distinct temporal and spatial scales, namely the conservation of an individual in the chemical domain and the evolution of a population in the historical domain.

**3.  Living is being, doing, developing, and evolving**

3.1  Living is information-processing? History repeats itself

Given this theoretical convergence of the two main traditions, and considering recent practical successes in implementing this new synthesis via synthetic biology [95], it may seem that the optimism pervading the field is well founded. The creation of all kinds of useful artificial life forms appears to be within our grasp, and the final mysteries of the origin and evolution of life on earth seem tantalizingly close to being resolved [35].

However, the confident promises of synthetic biology will sound all too familiar to those who know the history of 'synthetic psychology' – an approach better known as artificial intelligence (AI). Indeed, about half a century ago there was a similar optimism prevalent in the scientific community, stimulated by some early successes in robotics and machine learning, that the creation of artificial minds and conscious machines was just around the corner. Psychology, it was believed, was soon to be replaced by computer science. The driving force of that optimism, which in hindsight looks naïve and misguided, was a digital-information-centered theory of the mind that resonated with new advances in engineering and technology. Information technology has indeed turned out to be a great success in revolutionizing the industry, but the idea that cognitive science can be reduced to computer science is no longer in fashion. There are promising alternative theories of mind on offer, and it is clear that the role of symbolic AI will remain marginal [27]. On the contrary, the mind is essentially embodied, embedded, dynamical, and enactive [13, 28, 33, 102]. How ironic it is, then, that at the moment in which cognitive science is undergoing a major theoretical makeover toward a theory of the mind that is grounded in the whole organism [102], the science of life is at the same time extoling the virtues of trying to reduce the complexities of cellular biology to the level of 'logic circuits' [67] and 'computer programming' [3]. The history of science, it seems, is about to repeat itself. The reduction of biological life to computational logic is clearly not as straightforward as some recent advances in biotechnology may seem to indicate [29].

In particular, we note that, in a significant sense, the life as lived by the individual organism is still completely absent from the new consensus about the origin of life that we outlined above. On the one hand there is structural self-maintenance, and on the other hand there is informational self-replication. However, we know the former phenomenon from the general class of dissipative structures, and the latter phenomenon from the case of computer viruses – and neither of these two phenomena is typically considered as being alive. It is hoped that combining these two non-living phenomena will create a bridge to living phenomena, but we argue that this hope is likely unjustified unless the behavioral and developmental time scales are also taken into account. What the synthetic products of the information-compartment-metabolism-first approach are currently missing is the autonomous expression of goal-directed changes at the level of the individual organism as a whole. There is a complete lack of concern with the role of translational movement and transformational change, which could be studied in terms of traditional fields of biology including ethology and developmental biology.[2]

3.2 The timescales of life: Metabolism, behavior, development, and evolution

We propose that all of these aspects of life, i.e. metabolism, behavior, development, and evolution, form one unified phenomenon. Each of these processes, no matter its spatiotemporal scale, is integrated into one process of open-ended becoming. On this view, the possibility of distinguishing between them is simply due to the fact that they are expressed in terms of different timescales. Separating out one or more timescales to the exclusion of the others, e.g. historical change for the replicator-first approach or chemical rates for the metabolism-first approach, results in an impoverished and ultimately inadequate characterization of life. All known living beings are embedded within four broad interlinked categories of change:

---

[2] Although we do not focus on the role of morphology and ecology in this discussion, they are clearly important topics of future research. For example, in related studies it has been found that morphology is linked to self-motility in a model of autopoiesis [96] and environmental negative feedback is linked to self-individuation in a model of a reaction-diffusion system [107, Chapter 5.4].

Being: Metabolic events on this smallest timescale are taking place continuously in the physical and chemical domain of the organism. They are foundational in that they realize the concrete, spatiotemporally localized, existence of the individual living being in an autonomous manner via self-production [5].

Doing: Behaviors in this first intermediate timescale are unfolding in the relational domain of organism-environment interaction. The relational changes can be more or less tightly coupled to internal metabolic changes [18], but they are a non-reducible emergent property of the interaction process that cannot be conceptualized non-relationally.

Developing: Ontogenetic events on this second intermediate timescale are involved in making an individual become a structurally qualitatively different kind of individual within its own lifetime. Examples include some forms of learning, such as changes in neural connectivity, and morphogenesis.

Evolving: structurally qualitative changes in the historical lineage of generations of individuals take place on the longest timescales. Examples are code-based genetic, compositional genetic, and epigenetic forms of evolution that are shaped by natural selection, sexual selection, and/or natural drift.

Of course, the differentiation of the changes exhibited by living beings into these four distinct timescales should not be misunderstood in any absolute sense. Our starting point is to treat life as a unified phenomenon, and these distinctions do not reflect strict boundaries between the distinct timescales of becoming. While each of these timescales can be addressed in relative isolation, as demonstrated by their respective fields of study: biochemistry, ethology, developmental biology, and evolutionary biology, a complete understanding of life must be able to show how these different aspects are expressions of one and the same unified phenomenon[3]. They are mutually interdependent and yet non-reducible. We suggest that one way of moving biological theory forward is by introducing some intermediate timescales, namely behavior and development, into the current debates surrounding the origin of life. We need to consider that in the case of all extant living beings the 'self' referred to by the notions of *self*-maintenance and *self*-replication is a center of activity, i.e. an agent [85].

## 3.3 Toward a movement-first approach to the origin of life

It may be that the most minimal form of life that satisfies all of our timescale criteria would actually have to be a membrane-bound single-celled organism that is capable of code-based genetic evolution by means of natural selection. However, in what follows we will try to loosen some assumptions of the information-compartment-metabolism-first consensus in order to see how much can be achieved even with much simpler systems.

---

[3] There is an interesting analogy with robotics here, which has addressed behavior, some development, and evolution, but has so far failed to incorporate metabolism. Several significant AI problems arise from this specific shortcoming [32]. Similarly, the information-compartment-metabolism-first theory has addressed metabolism and evolution, but so far has failed to account for behavior and development. We therefore expect related problems to become apparent.

To begin with, we side with the metabolism-first approach in questioning the need for a genetic representational code at the origin of life. Relaxing the assumption of such a specialized information system seems reasonable. For instance, it has been argued that the 'RNA world' hypothesis faces considerable difficulties when confronted with realistic constraints of prebiotic Earth [92]. A promising alternative is to reject the requirement of a *code-based* genetic system at the beginning of evolution by natural selection. It is possible that genetic representation may not have been present at the origin of life, but appeared during later stages. For example, following the work of Segré, Lancet and colleagues [89-91], a protocell's chemical composition can itself serve as a 'compositional genome', which remains relatively well preserved during protocell division. Alternatively, it is possible that the composition's heredity is achieved through multiple attractors in the autocatalytic reaction network's dynamics [25]. Our argument is sympathetic with these non-representational approaches to the origin of heredity, but it is not dependent on them.

In addition, as discussed above, nowadays it is customary for both replicator-first and metabolism-first approaches to assume the existence of a physical boundary that protects either the RNA or the composition of the chemical mixture from adverse environmental influences, e.g. by assuming a lipid vesicle [54]. For instance, when this requirement of a boundary is combined with the idea of a compositional genome, we end up with the 'Lipid world' scenario of the origin of life [89]. However, by enclosing the RNA or other autocatalytic network in a relatively inert semipermeable membrane, which is often conceived as originating through external processes [15], these approaches are implicitly committed to the idea that the origin of life gave rise to behaviorally passive entities. It has been suggested that the preferential association of an ensemble of compounds "could be achieved by passive compartmentalization on the surface of fine particulate matter, within aerosol particles in the atmosphere, or within the pores of a rock" [45, p. 219]. And yet all of life as we know it today depends on an active process of organism-environment interaction and its adaptive regulation [5, 16, 63]. Far from serving merely as a *static compartment*, the membrane boundary around every living being is better conceived of as an *active interface* that is essential for behavior generation [8, 39]. It is precisely by means of this self-other interface that a cell regulates its metabolism and behavior through chemical and sensorimotor pathways.

The contrast between a static compartment and an active interface leaves us with two possibilities. On the one hand, we can continue to assume that life originally began enclosed in a static compartment and try to explain how this boundary later developed into an active interface. This idea derives from a long tradition of research [100], and it is still an integral part of the information-compartment-metabolism-first consensus. The challenge for this view is to solve the 'permeability problem' [15, 55]: how could chemicals be selectively exchanged across the compartment without the help of cellular proteins specialized for molecular transport? In modern cells, the membrane is covered with active pumps. Solving this problem was essential for cellular life because without an inflow of nutrients the cell will starve, without an outflow of waste it will poison itself, and without regulation of chemical concentrations an osmotic burst can rupture the compartment.

On the other hand, we can place priority on an active interface and adaptive behavior, and therefore relax the widespread assumption that a static compartment is needed for the first step in biological organization. It may seem that only a physical compartment can ensure the individuality of a living being as an entity that is distinct from its environment, but we argue that this is not necessarily the case. Even modern cells are quite porous when observed at the level of membrane pumps. What really matters is the operational coherence or closure of the living system as a whole. And although this operational closure may be enhanced through a self-produced physical boundary, such a spatial boundary by itself should not be confused with the organizational limits of the living system as one integrated system [108]. For example, in some origin of life scenarios, chemical interactions are sufficient for the self-maintenance of a coherent systemic identity [45]. As we will argue below, chemical gradients can serve this purpose, too. To be sure, such a flexible 'boundary' can make it more challenging for an individual to survive in unfavorable environmental conditions. However, it is also the case that some adverse effects of this vulnerability, both on the level of the individual and of the population, can be mitigated by benefits on the individual-level such as rapid multiplication and, especially, motility, adaptive behavior, and directed exploration – a possibility that has not yet been sufficiently considered. On this view, there is the possibility that compartments arose through a process of self-compartmentalization that was driven by interactive requirements.

The idea of motility at the origin of life appears to be far removed from the current concerns of the information-compartment-metabolism-first consensus. Nevertheless, as we will argue in the next section, it relates to several topical issues in relevant ways. Furthermore, it can draw philosophical support from an ongoing reevaluation of movement in the phenomenology of life [6, 93]. And, most importantly, we also know from existing artificial life research that some life-like behaviors can already be found in protocells and even in simple prebiotic chemistry. For instance, it has been shown that metabolic self-production can easily lead to spontaneous movement as well as adaptive gradient following, i.e. chemotaxis, in minimal models of protocells [18, 96]. This synergy between behavior and metabolism also plays an essential role for longer timescales; agent-based models of protocell evolution have shown that metabolism-based behavior can facilitate and guide evolution in several ways [19]. It has also been demonstrated that some of the chemicals typically favored for the synthesis of artificial protocells can spontaneously form oil droplets that exhibit self-sustained motility and a type of chemotaxis [37, 40, 103]. Accordingly, there have been calls for a new field of study in artificial life, variously labeled as 'homeodynamics' [42], 'chemo-ethology' [20], and 'chemical cognition' [39]. In the specific context of the science of the origin of life, we propose to call this newly developing perspective a *movement-first* approach.

In the following section we build on our previous models [29, 31, 107] in order to illustrate some of the issues related to a movement-first approach. The upshot is that, at least in evolutionary terms, it does not matter if compartmentless individuals are more prone to die from adverse changes in environmental conditions, as long as they are able to replicate quickly and move to different areas fast enough. It may therefore be the case that the evolution of a solid self-boundary was a secondary achievement that had to be balanced with the requirements of maintaining an active interface. Here we also see the importance of including a consideration of the intermediate timescales in which behavior takes place. Effectively, the population must be sufficiently distributed in its environment such that some of the individuals escape localized extinction events, but this distribution is most reliably achieved by self-movement of the individuals.

It is therefore conceivable that an early capacity for adaptive self-motility was already present at the origin of life. The model described in the next section is intended as a minimal proof of concept of this possibility. We emphasize that the aim of this model is not to provide a realistic simulation of prebiotic chemical conditions, nor do we claim that we have succeeded in modeling an actual living system. Rather, the aim is to advance our understanding of life by taking a closer look at the life-like properties of some of the 'transients' that exist somewhere between the inert and the living [24].

## 4. Toward a minimal model of the 'movement-first' approach

### 4.1 Precarious, individuated dissipative structures

Virgo [107] has argued that many of the properties of living organisms are shared by simple dissipative structures of the kind that form in reaction-diffusion systems. We review his reasoning below. Prigogine [73] coined the phrase 'dissipative structure' to denote a structure within a physical system that is actively maintained by a flow of energy and/or matter, rather than being an inert structure that is merely resistant to decay. Prigogine observed that living organisms are dissipative structures in this sense; however there are many other examples.

Given what has been argued in the previous section, a suitable starting point for our model of the movement-first approach to the origin of life would be a self-sustaining chemical processes that is a spatiotemporally coherent individual, and yet is non-compartmentalized. These criteria are met by a special class of dissipative structures, which Virgo ([107], Chap. 5) has called *precarious, individuated dissipative structures*. The class of living beings belongs to this class of structures. First, in addition to being a kind of dissipative structure, organisms have the property of being *precarious*, in the sense that if their structure is sufficiently disrupted it will stop being maintained (i.e. death). This emphasis on the precarious nature of the living goes against the popular idea that life is essentially about stability. The aim of stability would imply that the immortality enjoyed by non-living matter would be the ultimate yet unattainable goal of life. Instead, the process of living is better described as a transient with an end that can be postponed but not altogether avoided. Second, organisms are *individuated*, in the sense that they are spatiotemporally localized, and this localization is a result of the processes that make up the organism, rather than being imposed from outside. The notion that a living being is a precarious, individuated dissipative structure aligns our position with organism-centered approaches in biology and cognitive science, which are sometimes called 'enactive' (see also [4, 16, 17, 30, 43, 110]).

Virgo has pointed out that certain other dissipative structures share these properties with living organisms. One non-living example of this type is a hurricane [60]. It is dissipative in that it 'feeds' off a temperature gradient between the sea surface in the upper atmosphere; it is precarious in that if an important component is removed it can blow out (as will eventually occur if it passes over land); and it is individuated in that it is the cause of its own spatial localization. Of course, not all dissipative structures are precarious or individuated, and not all precarious, individuated dissipative structures share all properties of living systems. Nevertheless, studying this class of life-like structures provides a useful methodology for modeling some of life's basic properties, especially if they exhibit self-organized behavior [39, 107].

## 4.2 The Gray-Scott reaction-diffusion system

A simple and easy-to-study system that exhibits precarious, individuated dissipative structures is the Gray-Scott reaction-diffusion system, which was first studied in a two-dimensional context by Pearson [72]. This is a minimal model of chemical reactions taking place on a surface. The reaction modeled is a simple autocatalytic one, *A + 2B ➔ 3B*, meaning that when two molecules of *B* collide with one of *A*, they react to produce a third molecule of *B* while using up one *A* in the process. A second reaction, *B ➔ P*, represents the decay of the autocatalyst into an inert product that instantly leaves the system. The molecules *A* and *B* have a separate concentration at each point on a 2-D surface, represented by *a* and *b* (the concentration of *P* is not modeled). In addition, the 'food' molecule *A* is fed into every point at a rate proportional to 1 - *a*. This can be thought of as due to the surface being immersed in a solution of *A* at a constant concentration of 1.

Finally, in addition to reacting and being added to the system, the two chemical species can diffuse across the surface. Overall this gives rise to Equations 1 and 2

$$\frac{\partial a}{\partial t} = D_A \nabla^2 a - ab^2 + r(1-a) \tag{1}$$

$$\frac{\partial b}{\partial t} = D_B \nabla^2 b + ab^2 - kb \tag{2}$$

where concentrations *a* and *b* are functions of space as well as time, *r* and *k* are parameters determined by the rates of the two reactions and the feed process (the rate of the autocatalytic reaction has been set to 1 without loss of generality), and $D_A$ and $D_B$ are the rates at which the species diffuse across the surface. These equations can be solved numerically using a method that is akin to a cellular automaton, except that each cell contains a continually variable amount of the two chemical species.

Pearson observed that, depending on the choice of initial parameters, this system can form a variety of patterns, some of which are shown in Figure 1. Of particular interest are the spot patterns in Figure 1(f) and 1(g), since the spots have the properties of being individuated and precarious [107].

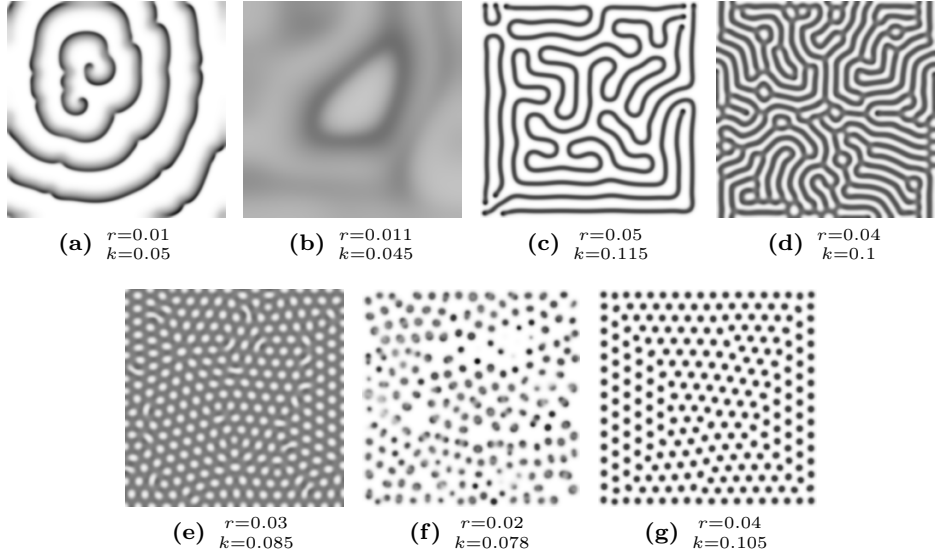

**Figure 1**: Examples showing the range of patterns exhibited by the Gray-Scott reaction-diffusion system with various parameters ($D_A$ = 2x10$^{-5}$ and $D_B$ = 10$^{-5}$ in each). The integration method and initial conditions are similar to those used by Pearson [72]. Patterns are chosen as exemplars of various phenomena; see Pearson [72] for a more systematic classification. (a) a spiral pattern; (b) a chaotic pattern of travelling waves; (c) a line pattern, whereby lines grow at the ends and then bend to fill space in a process reminiscent of a river meandering; (d) a labyrinth pattern; (e) a hole pattern; (f) a pattern of unstable spots, whose overall population is maintained by a balance between reproduction and natural disintegration; (g) A stable spot pattern, whereby spots reproduce to fill empty space and then slowly migrate into the more-or-less organized pattern shown (with a different choice of parameters, spots can be produced that are stable but cannot reproduce). (Figure taken from Virgo [107], p. 85)

We also know that many kinds of dissipative structures that are formed by reaction-diffusion systems are capable of sustained movement and even self-replication. This kind of self-organized motility has been investigated experimentally [50-52] as well as modeled mathematically [48, 72, 104]. The dynamics of self-replicating reaction-diffusion patterns have also been studied [80, 81]. In the dissipative structures of the Gray-Scott model we find cases of motility and replication as well, and this includes some kinds of individuated spots. We thus have all the basic requirements to begin our investigation of these 'spots' as a potential minimal model of the movement-first approach to life. We are particularly interested in whether the activity of this kind of reaction-diffusion system can be interpreted as taking place on the four timescales of metabolism, behavior, development, and evolution.

4.3 Metabolism

A reaction-diffusion spot can spontaneously emerge under appropriate conditions, and once it exists, it can self-maintain its precarious existence by means of a continuous turnover of chemical reactions (Fig. 2). As a self-producing network of chemical processes it satisfies the requirements of the first timescale of metabolism. It also provides the reference point of a spatiotemporal entity against which changes in other, slower timescales can be measured.

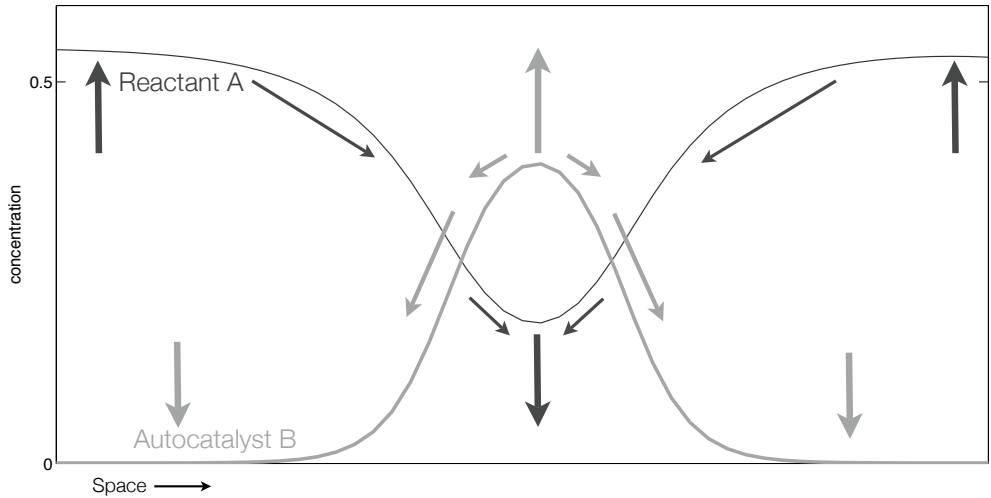

**Figure 2:** Concentration profile of a 'spot'; a precarious, individuated dissipative structure found in the Gray-Scott reaction-diffusion system. For clarity of display a one-dimensional version of the system is shown. The spatially individuated structure of the spot remains constant over time due to a precarious balance between reaction and diffusion processes. The arrows indicate the direction in which processes occur. For example, near the center of the spot, $B$ is continually produced by the autocatalytic reaction (upward arrow in the middle), but its concentration there remains constant due to the diffusive transport of $B$ towards the sides of the spot (angled arrows). This diffusion is in turn is balanced by the tendency for the autocatalyst to decay away in the regions where its concentration is low, i.e. outside the center of the spot (downward arrows). (Figure taken from Virgo [107], p. 86)

It is interesting to note in this regard that the spatiotemporal boundaries of a spot are intrinsically fuzzy. There is no distinct physical boundary. It is just as impossible to pinpoint the precise moment in time when the spot begins or ceases to exist, as the precise point in space where the spot ends and the environment begins. This is because the spot is a self-organizing phenomenon that is both continuous in time (temporal ambiguity) and continuous in space (spatial ambiguity). Nevertheless, at least an intuitive grasp of what constitutes an individual spot is possible; we either see an individual spot on the surface or we do not. The same fuzziness applies to the case of a living organism; for example, at the chemical level there is an inherent ambiguity of when and where a given molecule becomes a part of the organism. Similarly, it is possible that prebiotic compartments were very leaky, too [15]. This fuzziness is an advantage of the current model over other models that arbitrarily impose an absolute distinction between the inside and outside of a living system.

Once an individual spot has spontaneously formed, it will continue to exist even when it encounters a limited range of conditions that would not have enabled its original emergence. The fact that spots can exist outside of their original range of emergence is an indication that they are actively reproducing the viability conditions required for their existence, which is a strong criterion for autopoietic autonomy [30]. It shows that the complex system as a whole is more than a linear combination of its isolated parts by placing top-down constraints on the level of its components. Indeed, it is no different in the case of actual living beings: although the origin of life must have taken place in environmental conditions that favored the spontaneous emergence of one or more living beings, later living beings have had to actively contribute to the maintenance of their own conditions of existence in order to persist. However, living beings are different from other dissipative structures in that they adaptively regulate the self-maintenance of their conditions of viability, for example via behavior [7].

4.4 Behavior

For the purpose of this paper we define the concept of behavior broadly as a process in the relational individual-environment domain. The start of a behavior is induced by an instability or tension in that relationship, and a behavior ceases when that tension is resolved or transformed into a different kind of tension, which elicits a different kind of behavior [17]. We take the view that all behavior is characterized by an essential agent-environment asymmetry, which is centered on the self-producing activity of the individual [4]. The tension that triggers a behavior may originate in the environment, but the fact that there can be such a tension at all is an achievement of the self-constitution of the individual, which brings about the relational domain in the first place. In this sense the metabolic activity of the individual is the ultimate source of all its behavior, and this behavior arguably acquires its meaning for the point of view of the individual via this direct metabolic grounding [32].

The general term 'behavior' covers a huge variety of changes in all kinds of entity-environment relations, so some distinctions are in order. One important distinction in biology and psychology is between *reactive behavior*, namely behavior that is directly triggered by events in the environment, and what we call *intrinsic behavior*, namely behavior that is spontaneously performed by the individual. In what follows we are particularly interested in intrinsic movement, i.e. self-movement, but the distinction between reactive and intrinsic is not an absolute one. On the one hand, all biological systems have internal state and their 'reactive' behavior is therefore always also a function of their history, and, on the other hand, expression of intrinsically generated behavior always takes place within the context of external conditions and events. Nevertheless, it is useful to consider that a behavior can be more or less driven by internal and external factors. This is true for living beings and also, we argue, for the reaction-diffusion spots.

To be clear, we are not claiming that the behavior of spots shares the same underlying mechanisms with the behavior of living beings. While the former kind of behavior is a side-product of metabolic activity, the latter is based on specialized and metabolically semi-detached regulatory processes [5], although metabolism-dependent behaviors may be present in some single-cell organisms [18]. In any case, even if the behavior of the reaction-diffusion spots does not satisfy strict organizational criteria for autonomous behavior or agency, for instance because the internal processes do not exhibit differentiated functions [66], it is important for the science of the origin of life to acknowledge that behaviors can be generated even by proto-life systems using much simpler mechanisms.

<u>Reactive behavior.</u> The spots exhibit a clear type of reactive behavior with respect to differences in chemical gradients in their surroundings. We can describe this behavior in the biological terms of approach and avoidance: the spots are capable of following chemical gradients that increase the concentration of their constituents, which is akin to bacterial chemotaxis, and they are also capable of avoiding chemical gradients that decrease the concentration of their constituents. This gradient-based movement is a minimal example of metabolism-dependent behavior, which provides the spots with a basic capacity to adapt to environmental changes. For example, we can attract the spots by 'feeding' them, i.e. by using a 'virtual pipette' to add constituents into their vicinity. If we feed too much food to a spot, then it will divide and form two spots. On the other hand, when we remove constituents from a nearby spot by using the pipette, the spot will tend to move away from the pipette. In this way it is possible to chase spots around the simulated surface. If the pipette is too fast and gets too close to a spot, it destabilizes the metabolic activity of the spot in such a way that the spot is no longer sustainable and dies.

If there are several spots in the environment, then these approach and avoidance behaviors will make them interact in certain ways. This is because a spot consumes the food in its proximity, thereby surrounding itself with a negative gradient that keeps other spots away. If the spots did not tend to move away from one another then they would merge rather than remaining separate; these approach and avoidance behaviors therefore form an important part of the individuation process.

Again, it is important to note that although these behaviors are reactive in the sense that they only occur in the presence of an appropriate environmental trigger, they are not reactive like the responses of a stateless system. On the contrary, they are the result of a continually active growth process. The spot moves in an adaptive manner because the autocatalyst grows faster on the side where the food concentration is higher. Therefore, even though these behaviors appear as reactive in the behavioral domain, they are nevertheless, like all behaviors, active in the metabolic domain.

<u>Intrinsic behavior.</u> However, in order for the spot to move around spontaneously, i.e. even in the absence of environmental triggers, it must create its own instabilities that trigger the appropriate responses. Of course, because it is a dissipative structure, the spot as a whole is already in a far-from-equilibrium state. But since the instability that is produced by the reaction-diffusion system is symmetrical in all directions, the spot remains stationary without any environmental gradients. Accordingly, a self-moving spot requires a means of self-producing an asymmetrical distribution in the domain of individual-environment relationships. For example, in the case of a self-propelling oil-droplet, symmetry breaking related to the expulsion of waste products was found to be essential for the generation of movement [37].

We modified the model of the Gray-Scott reaction-diffusion system by implementing the concentration of the waste product P. We tested a number of parameter settings for achieving a similar kind of spontaneous motility for the spots. To do this we replaced Equations 1 and 2 with the following set of Equations 3-5:

$$\frac{\partial a}{\partial t} = D_A \nabla^2 a - e^{-wp} ab^2 + r(1 - a) \qquad (3)$$

$$\frac{\partial b}{\partial t} = D_B \nabla^2 b + e^{-wp} ab^2 - kb \qquad (4)$$

$$\frac{\partial p}{\partial t} = kb - k_p p \qquad (5)$$

A new equation has been added to the reaction-diffusion system, modeling the spatial distribution of P. The component P is created when B decays, and P itself decays at a very slow rate (we use $k_p = 0.0002$). The concentration of P modulates the kinetics of the *A + 2B* ➔ *3B* reaction: its forward reaction rate is now given by $e^{-wp}$, so that a buildup of P at a particular point will inhibit the autocatalysis of B. Figure 3 shows an example of the dynamics of this new system, with the parameters $w = 0.015$, $r = 0.032$ and $k = 0.0942$, integrated with a time step of 0.5 units.

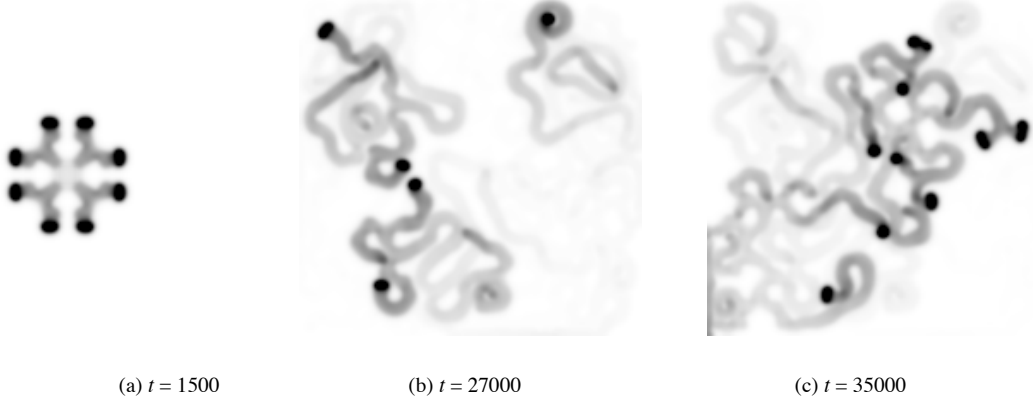

(a) $t = 1500$     (b) $t = 27000$     (c) $t = 35000$

**Figure 3**: A snapshot of the dynamics of the system defined by Equations 3-5. With this version of the equations the spots create a build-up of waste product that is detrimental to them, and this causes them to move away from it, leaving a trail behind them. The colors have been adjusted so that the spots of autocatalyst B appear black, whereas the waste product P appears as a lighter shade of gray. (a) The system is seeded with an initial square of autocatalyst; the figure shows the system shortly afterwards. The initial conditions must be chosen so that the autocatalyst does not immediately use up all the nearby nutrients, because the spot would then decay toward death. But the conditions can otherwise be chosen arbitrarily, since they make little difference to the final behavior. (b) Note that the spot towards the upper right has followed a spiral path, accidentally blocking itself into an area with a high waste concentration. Shortly after this (within 1000 time units) it ceased to persist; the concentration of the autocatalyst dropped below the threshold needed for self-maintenance and the spot disappeared. (c) More spots have appeared again. The overall dynamics of this system are complex, with the size of the population of spots tending to oscillate as the overall amount of waste in the system builds up and then slowly decays.

The inhibitive effect of waste P causes the spots' overall movement, as long as there is an asymmetrical distribution of P around the autocatalyst. If there is a gradient of P across a spot, then the growth rate of B will be higher on the side that has less P. This causes movement over time, as the side with less growth (due to more P) dies away more rapidly, thereby producing more P. As the spot moves it leaves a trail of P behind it, which in turn maintains the concentration gradient of P across the spot.

It could be argued that the waste-driven movement is still more externally driven than internally driven. However, there is another way of achieving spontaneous motion of the spots. This is achieved by modifying the original Gray-Scott reaction-diffusion system by introducing a second autocatalyst to the system, which feeds not on the 'food' molecule but on the other autocatalyst [107]. That is, the reactions $B + 2C \rightarrow 3C$ and $C \rightarrow P$ are added to the system, so that Equations 1 and 2 are extended to Equations 6-8, where $D_C$ is the rate of diffusion of C, and $k_1$, $k_2$ and $k_3$ are the rate constants for the reactions $B \rightarrow P$, $B + 2C \rightarrow 3C$ and $C \rightarrow P$, respectively.

$$\frac{\partial a}{\partial t} = D_A \nabla^2 a - ab^2 + r(1-a) \tag{6}$$

$$\frac{\partial b}{\partial t} = D_B \nabla^2 b + ab^2 - k_1 b - k_2 bc^2 \tag{7}$$

$$\frac{\partial c}{\partial t} = D_C \nabla^2 c + k_2 bc^2 - k_3 c \tag{8}$$

With an appropriate choice of parameters, the effect of this modification is to produce the usual spots of primary autocatalyst, but this time accompanied by a small region of the secondary autocatalyst. Since the secondary autocatalyst feeds on the primary one, the spot of primary autocatalyst tends to avoid it by moving away, while the secondary spot follows. This gives the secondary autocatalyst the appearance of being attached as a 'tail' behind the primary spot (Figure 4). The spot-tail system as a whole moves around spontaneously even in a homogeneous environment. In the sense that this self-motility depends on the internal constitution of the whole spot-tail system itself, we can characterize it as a form of intrinsic, non-reactive behavior. Related work has also found the emergence of spatial patterns with movement and dynamics of their own, for example spiral waves [41]. This model is different in that it shows the emergence of spatially distinct individuals that are self-moving.

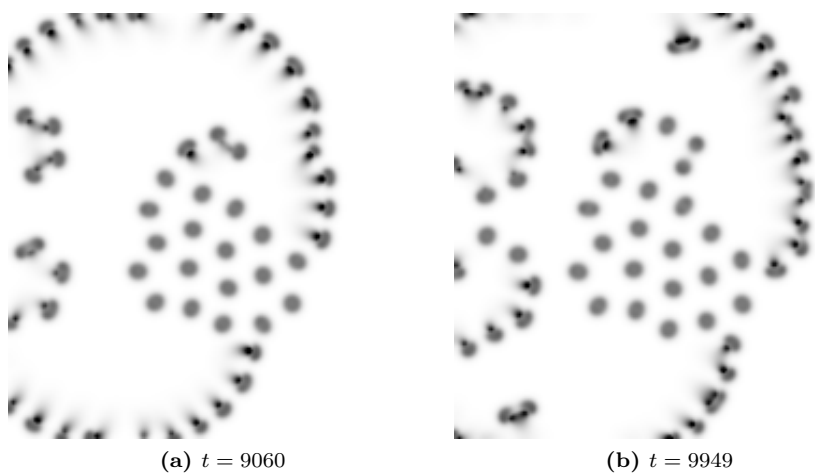

(a) $t = 9060$      (b) $t = 9949$

**Figure 4**: Two snapshots of the system resulting from Equations 6-8, integrated on a surface of 2 by 2 units, with the parameters $D_A = 2 \times 10^{-5}$, $D_B = 10^{-5}$, $D_C = 10^{-6}$, $r = 0.0347$, $k_1 = 0.2$, $k_2 = 0.8$ and $k_3 = 0.005$. The colors are adjusted so that the secondary autocatalyst $C$ appears as a darker shade of gray than the primary autocatalyst $B$. A group of spots with tails can be seen on the mid-left side of plot (a), and after duplication in plot (b) in the same place. Some tail-less spots can be seen as well, their tails having been lost in the process (hence, this is limited heredity with variation). The spots with tails move constantly in the direction facing away from their tails at a rate of approximately $4 \times 10^{-4}$ distance units per time unit, which results in their colonizing the empty part of space more rapidly than the tail-less spots. However, with this choice of parameters, the tailed spots cannot invade areas occupied by tail-less spots, and they are eventually crowded out and become extinct. (Figure taken from Virgo [107], p. 105).

Although the spot-tail system is not strictly speaking an autocatalytic 'hypercycle' [21], because the direct chemical dependency between the two catalysts is not mutual, it nevertheless can be considered as 'symbiotic' to some extent [29, 49]. More specifically, although the tail is parasitic on the primary spot (since it contributes nothing to it metabolically), their jointly induced movements in the system-environment domain can be adaptive in some environments. For instance, with certain parameter settings of the simulation, the spot-tail systems can reproduce more rapidly than the spots without tails, and their movement also tends to make them colonize new areas more rapidly. Given the selective advantage introduced by the behavior of the system as a whole, we have elsewhere introduced the notion of a 'behavior-based hypercycle' [29]. Figure 5 shows an example of a scenario where, over longer timescales, spot-tail systems are better adapted than tail-less spots. This is due to the 'parasite'-enabled exploratory behavior, which helps to prevent localized extinction events from killing the whole population.

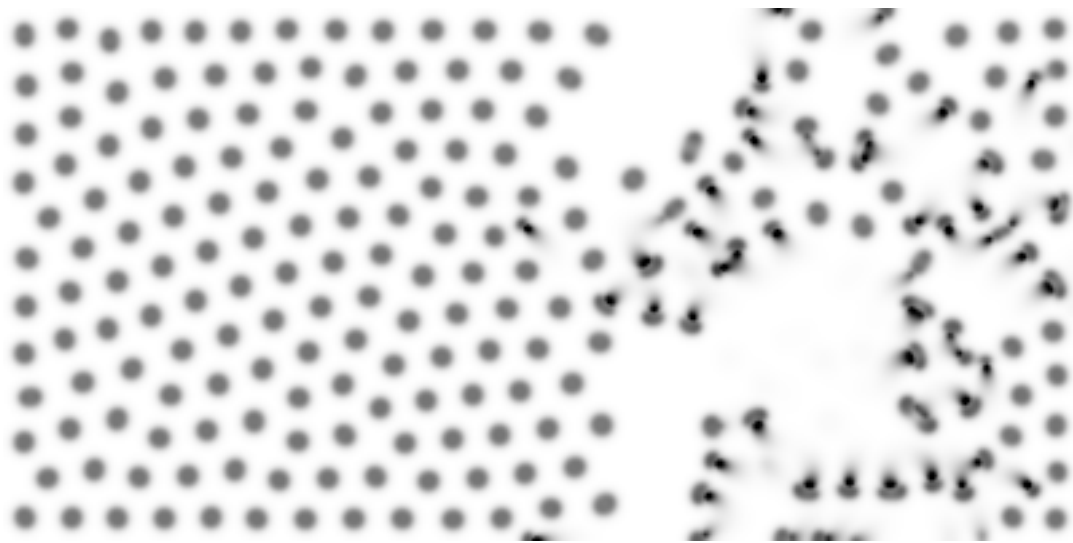

**Figure 5**: A snapshot from the same system shown in **Figure 4**, with the same parameters, except that randomly chosen areas in the right-hand side of the surface are occasionally cleared by an externally induced cataclysm (i.e. the food concentration in a random 0.5-by-0.5 area is set to zero every 1000 time units). The spots with tails are able to persist in this right-hand side due to their ability to colonize the cleared areas more rapidly than the spots without tails. But in the left-hand side of the figure they are out-competed. (Figure taken from Virgo [107], p. 105)

The replicator-first approach to the origin of life claims that parasitic reactions are a significant problem for the metabolism-first approach, because of their unavoidable detrimental metabolic effects (and hence, another reason for the necessity of a compartment [99]). Positive effects of parasites have not been considered so far [29]. Although it has been previously observed that the addition of a parasite can induce spatial self-structuring [87], most research has focused on how to mitigate or even expel parasites altogether [9]. It may seem that symbiosis with parasites is dependent on their direct chemical contribution [47]. However, as our reaction-diffusion model has demonstrated, this is not always necessarily the case, especially considering the role of events taking place at intermediate timescales [29]. The important point is that *what may be viewed as detrimental on the metabolic timescale (i.e. a 'parasitic' reaction) can induce novelty on the behavioral timescale (i.e. self-motility), which then turns out to be adaptive on the evolutionary timescale (i.e. faster replication and wider population distribution)*. Accordingly, given that most chemical studies of the origin of life completely ignore the possibility of behavioral effects at the system level having a differential effect for natural selection, they may be missing essential aspects of the phenomenon.

The model of the spot-tail system is important in another respect. Several researchers have emphasized that simple dissipative structures are not suitable starting points for understanding the origin of life, in particular because such structures generally lack distinct functions based on organizational differentiation [64, 66, 86]. It appears that all local processes of simple dissipative structures are subsumed under the single intrinsic goal of global self-maintenance. In other words, "it is an 'all or nothing' situation: a compound reacts or does not react. Its absence may destroy the global pattern, but it does not modulate or shift it in any specific way" ([64], p. 593). But to have distinct functions requires organizational differentiation, and "a self-maintaining system is organizationally differentiated if it produces different and localizable patterns or structures, each making a specific contribution to the conditions of existence of the whole organization" ([66], p. 826).

We suggest that the spot-tail is an example of a dissipative structure that can satisfy these requirements. First, its self-motility is a behavior-based contribution to global self-maintenance under some conditions, thereby integrating the 'tail' into the system's overall organizational closure [29]. Second, the ability of self-motility is generated by a distinct function, which is dependent on the contribution of a localizable structure, namely the 'tail' of the 'spot'. Third, the 'tail' can be removed without destroying the original 'spot', thereby demonstrating that this is not simply an 'all or nothing' situation. Instead, the spot-tail structure exhibits a kind of hierarchical decoupling of the tail from the main spot, because the latter can exist without the former but not vice versa. What this analysis shows is that the necessary requirements for the origin of life, as currently envisioned by these authors, are in fact realizable by more minimal systems than they expected.

4.5 Development

We conceive of the notion of development in a broad way so as to include any structural changes of the organism, which turn it into a qualitatively different kind of being in its own lifetime. These structural changes can include (ordered in roughly increasing temporal scale) growth, habituation, learning, and ontogeny. Not all forms of life exhibit all of these variations of development to the same extent, but all display some capacity for structural transformation.

In a very minimal sense, development can already be seen in single spots. When they exhibit directional movement, they do so because of ongoing structural changes: they grow toward the increasing gradient, and decay away on the other side. They are like plants in that growth and behavior are not always readily separable. In addition, we find more complex lifetime structural changes. These changes typically proceed via the incorporation of external elements rather than via internally generated changes. For example, a spot can be turned into a spot-tail system by being 'infected' with a tail from a passing spot-tail system. During this encounter the chemical species of the other's tail is incorporated into the functional organization of the spot. The emergence of a spot-tail system is therefore perhaps akin to a minimal prebiotic example of 'consortia', i.e. a tight coupling between two or more different microorganisms that associate during growth to form characteristically ordered structures. Consortia has been mentioned in the context of the hydrothermal vent scenario of the origin of life, e.g. a syntrophic metabolic relationship between $CH_4$-oxidizing archaea and sulphate-reducing bacteria [57]. In addition to the emergence of a spot-tail system Virgo [107] observed a second, related kind of process in another reaction-diffusion system (with a different set of equations), whereby two nearby spots consisting of mutually complementary catalytic reactions join together to form a multi-spot system, thus forming a proper chemical hypercycle. Future work could consider whether there is the possibility of prebiotic endosymbiosis.

## 4.6 Evolution

We have already observed that there is a heritable difference between a spot with a tail and a spot without tail. Once a spot acquires a tail due to a lifetime event, there is a good possibility that this feature will be passed on to the next generation when the spot-tail multiplies through division (see Figure 4). This evolution by infection is not as odd as it may seem at first, because symbiogenesis has likely played a significant role in the evolution of modern cells as well [56]. However, this spot-tail system is still lacking a decoupled genetic system with which to encode such differences in a relatively metabolism-independent manner. In our analysis of the spots' evolutionary capacity we therefore only focus on epigenetic evolution and on evolution by natural selection with a compositional genome.

<u>Epigenetic evolution.</u> One of the main epigenetic factors of inheritance is the specific time-space configuration in which an individual is born. A famous case is the beaver's dam, which, once constructed, provides a home for subsequent generations. This kind of inheritance can also occur in the case of reaction-diffusion spots. For instance, the offspring of those spots, which happened to divide because of a high concentration of nutrients, will also find themselves in a situation with high concentration of nutrients. Of course, the situation is no different for other examples of protocells, which divide with sufficient nutrients. But the spots have an additional advantage in that they tend to seek out areas with conditions that are more favorable, a behavior which not only ensures more chances of replication but also a better start for their offspring.

<u>Composition-genomic evolution.</u> The current chemical composition of spot can be considered as both its phenotype and genotype combined. The idea is that this kind of 'compositional genome' could have enabled protocellular evolution by means of natural selection even in the absence of any specialized information-carrying component such as RNA and DNA [91]. For instance, spots with tails can be seen as undergoing a Lamarckian form of inheritance, whereby traits that have been acquired during an individual's lifetime are passed along to the offspring. Once a spot has acquired a tail (perhaps by passing near to another tailed spot), it will divide in a way that often, but not always, results in offspring that have tails.

We also find a difference in selective pressure since in some environments the spots with tails are more viable than the single spots on their own (see Figure 5). This is because spot-tail systems move around even in the absence of chemical gradients, and they are thereby able to minimize the impact of localized catastrophic events. Greater spatial distribution lessens the risk of the population going extinct. In this scenario the original single-spot constituents may therefore die out eventually, while the spot-tail variant persists. Here we therefore have all the necessary elements of evolution by natural selection as it is standardly conceived, namely reproduction, variance, and selection, but with limited rather than 'unlimited heredity' [97]. A challenge for future work is to find a way in which such simple dissipative structures can regulate their self-production in a mediated manner, such that the generational transmission of the mediating structures itself can give rise to a more open-ended kind of heredity.

## 5. Discussion

The model has served as a proof of concept that even simple reaction-diffusion spots can exhibit many essential life-like characteristics, where life is conceived as a unified process of being, doing, developing and evolving. We have focused on the importance of self-organized motility and behavior in the context of current debates on the origin of life. In this discussion we would like to draw attention to the shortcomings of the current model, and to consider possible ways of overcoming them.

The spots satisfied the basic requirements of metabolism (self-creation) and behavior (self-motility). In fact, they are even capable of adaptive behavior that resembles the foraging behavior of actual bacteria (nutrient gradient following). The spots are also capable of some proto-development through the incorporation of additional external elements, and these lifetime changes are inheritable over generations. Taken together these findings suggest that the spots meet the criteria of undergoing changes within the four major timescales that are characteristic of life, namely metabolism, behavior, development, and evolution.

But are these spots really a model of the phenomenon of life? Life is not only expressed across those four major timescales, it is expressed in a characteristically *open-ended* manner of becoming [31]. No matter what timescale, life is always ready to surprise us. And it is precisely with regard to this capacity for open-endedness that the limitations of the current model are apparent. How far can the reaction-diffusion approach be scaled up? Is it possible to set up the environmental conditions such that a more complex network of dissipative structures emerges? Would this network be capable of new useful behaviors? By which mechanism could such a network learn to improve its behaviors? How could it reproduce itself reliably? And by which kinds of methods could we observe such a complex chemical system?

An important issue in sorting out these questions is the problem of mediation. At the moment these structures are only capable of what has been called "metabolism-based" behaviors [19]. Although the spot-tail system exhibited a minimal form of hierarchical decoupling, more complex behaviors require a more systematic mediation between the constitutive (metabolic) and interactive (behavioral) domains. Indeed, the whole history of life on earth can be interpreted as the evolution of increasingly complex forms of mediation that enabled the emergence of new forms of autonomy [44].

A related issue that still needs to be tackled in future models of this kind is how to introduce the possibility of solid structures. In the current model the spots are fully transparent to environmental interactions, although chemical gradients may constitute some boundaries. This extreme chemical transparency effectively turns the whole spot into an interface with its environment. In order to enable a more open-ended increase in complexity, it may eventually become necessary for these systems to localize their interfaces at their spatial boundaries. In other words, one important form of mediation is a sensorimotor surface. In addition, internal differentiation between the constitutive elements that are responsible for self-creation and those that are needed for regulating interaction leads to increased behavioral autonomy [5]. This differentiation may also enable further specialization of these elements for their respective tasks, since they no longer need to do both tasks at the same time.

Given the continuing dispute about the nature and role of the genetic system in the information-compartment-metabolism-first consensus, it is of interest to determine to what extent a representational genetic code is in fact necessary for the phenomenon of life, as we have characterized it. One practical way to address this issue, which we have pursued in this paper, is to determine how life-like a dissipative structure can be (and become) without having any dedicated genetic code that could map between a genotype and a phenotype. But if we do not assume that such a representational code already existed at the origin of life, then we are faced with the problem of how a distinct genotype could have arisen during the history of life.

We speculate that the genetic system may have originally had a function on the level of intermediate timescales of behavior and development. Clearly, in order for the genetic system to be retained and selected on the generational timescale of evolution, it must have already had a useful lifetime function. For instance, it could have started as a way of mediating processes of self-construction, which would have facilitated the adaptive regulation of behavior and developmental reorganization of metabolic and interactive pathways. It could also have served as a way facilitating the learning and memory capacities of the individuals. Given that these complex functions must have already been heritable if they were originated (or at least optimized) by evolution via natural selection, then their potential for exaptation in the service of a dedicated genotype may have been co-selected from the start.

**Acknowledgments**

Froese is funded by the Japanese Society for the Promotion of Science (JSPS), and his research is financially supported by their Grant-in-Aid program. This paper benefitted from the helpful comments of three anonymous reviewers. We would also like to thank the organizers, reviewers, and attendees of ECAL'11, where an earlier version of this paper was presented [31].